\documentclass[journal]{IEEEtran}
\usepackage{amsmath,amsfonts}
\usepackage{algorithmic}
\usepackage{array}
\usepackage[caption=false,font=normalsize,labelfont=sf,textfont=sf]{subfig}
\usepackage{textcomp}
\usepackage{stfloats}
\usepackage{url}
\usepackage{verbatim}
\usepackage{graphicx}

\usepackage[numbers,sort&compress]{natbib}
\usepackage[table]{xcolor} 
\usepackage{booktabs}
\usepackage{multirow}
\usepackage{balance}
\usepackage{pifont}
\usepackage[pagebackref=true,breaklinks=true,letterpaper=true,colorlinks,bookmarks=false]{hyperref}

\def\BibTeX{{\rm B\kern-.05em{\sc i\kern-.025em b}\kern-.08em
    T\kern-.1667em\lower.7ex\hbox{E}\kern-.125emX}}

\begin{document}
\title{DFR-Net: Density Feature Refinement Network for Image Dehazing Utilizing Haze Density Difference}

\author{
Zhongze Wang, Haitao Zhao, Lujian Yao, Jingchao Peng, Kaijie Zhao.
\thanks{
Manuscript received XXXX 00, 0000; accepted XXXX 00, 0000. 
Date of publication XXXX 00, 0000; date of current version XXXX 00, 0000. 
The associate editor coordinating the review of this manuscript and approving it for publication was XXXX.
(Corresponding authors: Haitao Zhao.)
 
Zhongze Wang, Haitao Zhao, Lujian Yao, Jingchao Peng, Kaijie Zhao are with East China University of Science and Technology, Shanghai 200237, China
(e-mail: htzhao@ecust.edu.cn)
}
}

\markboth{Journal of \LaTeX\ Class Files,~Vol.~18, No.~9, September~2020}%
{How to Use the IEEEtran \LaTeX \ Templates}

\maketitle

\newcommand{\gr}{\rowcolor[gray]{.95}}
\newcommand{\rt}{\textcolor[rgb]{0.75,0.25,0.25}}
\newcommand{\bt}{\textcolor[rgb]{0.25,0.25,0.75}}

\begin{abstract}
    In image dehazing task, haze density is a key feature and affects the performance of dehazing methods. However, some of the existing methods lack a comparative image to measure densities, and others create intermediate results but lack the exploitation of their density differences, which can facilitate perception of density. To address these deficiencies, we propose a density-aware dehazing method named Density Feature Refinement Network (DFR-Net) that extracts haze density features from density differences and leverages density differences to refine density features. In DFR-Net, we first generate a proposal image that has lower overall density than the hazy input, bringing in global density differences. Additionally, the dehazing residual of the proposal image reflects the level of dehazing performance and provides local density differences that indicate localized hard dehazing or high density areas. Subsequently, we introduce a Global Branch (GB) and a Local Branch (LB) to achieve density-awareness. In GB, we use Siamese networks for feature extraction of hazy inputs and proposal images, and we propose a Global Density Feature Refinement (GDFR) module that can refine features by pushing features with different global densities further away. In LB, we explore local density features from the dehazing residuals between hazy inputs and proposal images and introduce an Intermediate Dehazing Residual Feedforward (IDRF) module to update local features and pull them closer to clear image features. Sufficient experiments demonstrate that the proposed method achieves results beyond the state-of-the-art methods on various datasets.
\end{abstract}



\section{Introduction}

\IEEEPARstart{H}{aze}
 is a common atmospheric phenomenon caused by the accumulation of aerosol particles. It can cause severe quality degradation of images, which can affect subsequent computer vision tasks. Therefore, developing effective techniques for haze removal is essential to improve the quality of images and ensure accurate results of downstream tasks \cite{lin2022msaff,li2019pdr, liu2021joint}. 

After decades of study, researchers \cite{drake1985mie,mccartney1976optics} model this atmospheric phenomenon as:
\begin{equation}\label{eq:1}
    \textbf{I}(x)=\textbf{J}(x)t(x)+\textbf{A}(1-t(x))
\end{equation}
where $\textbf{I}(x)$ represents the hazy image; $\textbf{J}(x)$ represents the clear image; $t(x)$ and $\textbf{A}$ stand for the transmission map (T-map) and the global atmospheric light separately. And this model is commonly called the atmospheric scattering model (ASM).

\begin{figure}[t]
  \centering
  \includegraphics[width=\linewidth]{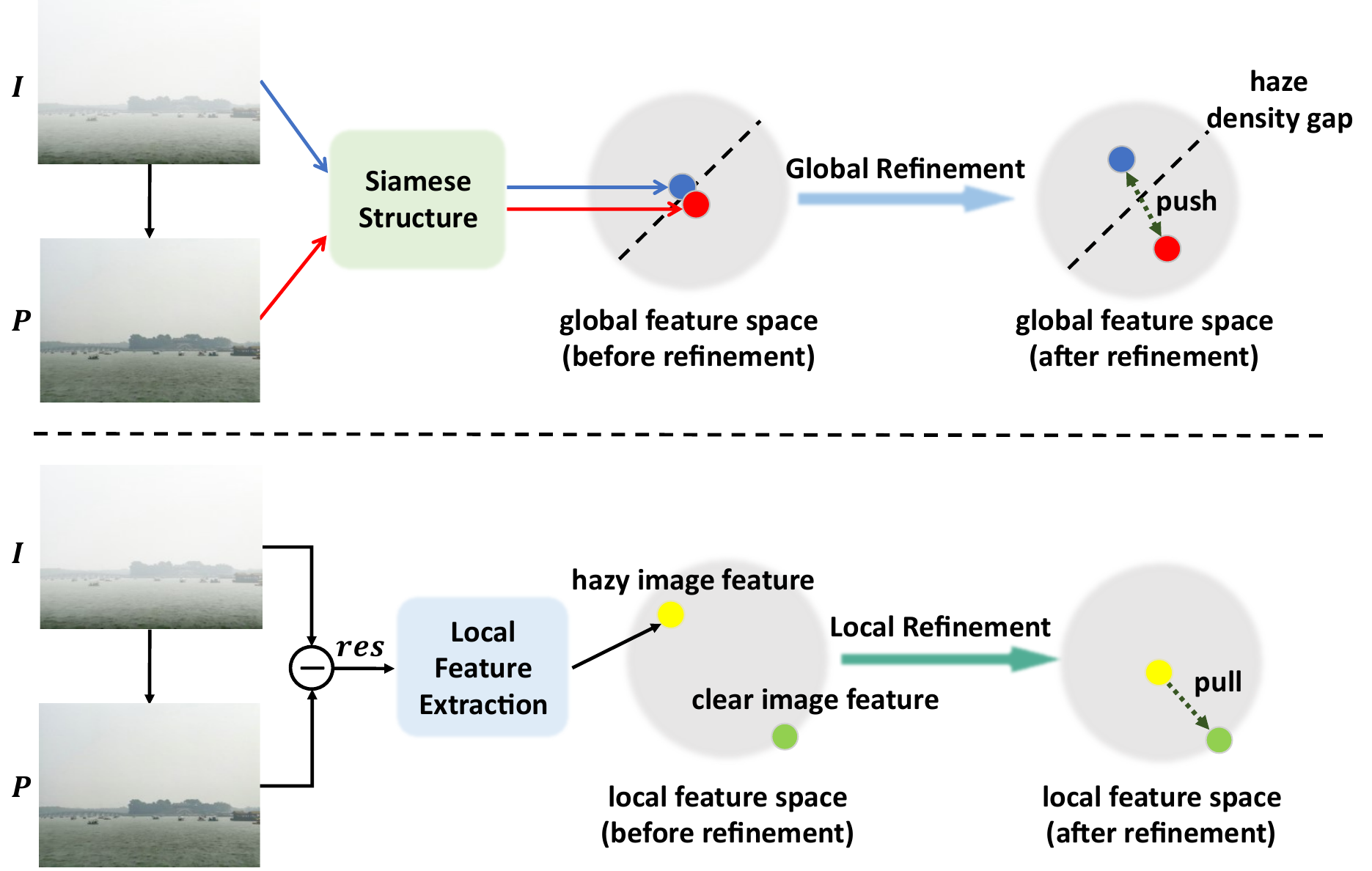}
  \caption{The main idea of our DFR-Net. By utilizing the density difference between the input (I) and the generated proposal image (P), global and local density features are refined in different ways: features of different global densities are pushed farther apart and local density features are pulled in towards clear image features.}
  
  \label{fig:head}
\end{figure}


To improve image quality and highlight image details captured in hazy weather, numerous image dehazing methods have been proposed. Prior-based methods \cite{he2010single, zhu2015fast, fattal2014dehazing, berman2016non, berman2018single} rely on statistical analysis of haze images and handcrafted priors to recover haze-free images. However, these methods have limitations in their robustness due to their reliance on specific assumptions, which may not hold in different scenes. 

With the success of deep neural networks in high-level tasks, data-driven dehazing methods \cite{ren2016single, li2017aod, berman2018single, chen2019gated, wu2021contrastive, yeperceiving} have become mainstream. Compared to prior-based methods, deep learning methods demonstrate stronger capabilities in feature extraction and image restoration. However, early deep learning methods neglect the uneven distribution of haze, resulting in redundancy in network design and inefficiency in feature extraction \cite{yeperceiving}. One idea to improve these methods is to enable the network to learn features about the haze density.

Research works \cite{yeperceiving,zhang2021hierarchical,deng2020hardgan,yang2022self,lou2020integrating} have proposed density-aware dehazing methods. Haze density describes the distribution of haze and impacts the effectiveness of a dehazing method. Several methods \cite{zhang2021hierarchical, yang2022self, lou2020integrating} estimate the T-map to obtain haze density information, which is inversely proportional to the T-map \cite{zhang2021hierarchical}. However, these methods require T-map labels, which can be difficult to obtain. To avoid this problem, methods \cite{deng2020hardgan, yeperceiving} directly extract density-related features. Nonetheless, these methods still have shortcomings: they lack a comparator to measure density, and the learning process of density features lacks interpretability.

As image dehazing is an ill-posed problem, estimating a clear image directly from a hazy input using a feedforward network is challenging \cite{bai2022self}. Additionally, distinguishing haze densities from a single hazy image is also difficult, and a comparable image that helps the network learn to perceive different haze densities is necessary for density-awareness. While some research works \cite{bai2022self,yeperceiving,chen2021desmokenet} have proposed methods that generate intermediate images to alleviate this problem, they do not take the density differences between the intermediate results and the hazy inputs into much consideration, limiting the network's ability to fully perceive haze density.

To address this issue, we propose a Density Feature Refinement Network (DFR-Net) that achieves density awareness by utilizing density difference. To fully explore haze density feature, we first generate a dehazing proposal image ($\textbf{P}$, as shown in Fig. \ref{fig:head}), that can provide a comparison of density information from both global and local perspectives. Our proposal image is generated by a simple U-net structure and has an overall lower haze density than the haze input ($\textbf{I}$), which brings global density differences. The different dehazing performance of different areas reflected in the dehazing residuals ($\textbf{res}$) leads to differences in local density and hints to areas with haze removal challenges. Subsequently, we propose a 2-branch structure consisting Global Branch (GB) and Local Branch (LB) \emph{based on the density difference information between} \emph{\textbf{P}} \emph{and} \emph{\textbf{I}} to comprehensively extract density features and refine them in different ways.

In detail, the scenes are the same between $\textbf{P}$ and $\textbf{I}$, differing only in haze density. Therefore, we propose to use Siamese structures for feature extraction for $\textbf{P}$ and $\textbf{I}$, enabling the network to understand both the different densities in one forward process. In addition, considering that the two features contain part of the same information, to push them farther away and highlight the density information, we design a Global Density Feature Refinement (GDFR) module to refine the features. Locally, the dehazing residual between $\textbf{P}$ and $\textbf{I}$ contains hints of the local density information. The dehazing residual represents the performance of dehazing and areas with small residual values tend to be more heavy or hard dehazing regions. Hence, we propose to learn local density features from the dehazing residual ($\textbf{P}-\textbf{I}$) in a split and merge (S\&M) way. To refine local features, several Intermediate Dehazing Residual Feedforward (IDRF) modules are used, which can pull local features closer to clear image features. The refinement illustration is exemplified in Fig.\ref{fig:head}. Both the two branches give a predicted haze-free image and we perform an adaptive fusion on them to gain the final dehazing result. 

Compared to other density-aware methods, on the one hand, our DFR-Net does not rely on additional T-map annotations and does not predict the T-map, thus reducing the manual workload and the possible loss of information due to the T-map prediction process \cite{jin2023dnf}. On the other hand, unlike the way other methods extract density features, e.g., PMNet extracts density features from the splicing of hazy inputs and pseudo-haze-free images via an SHA module, we analyze the relationship of haze density between $\textbf{P}$ and $\textbf{I}$, and design a network with interpretability to extract and refine density features from the density difference. This makes our DFR-Net contain density-related prior knowledge.

Overall, the main contributions of our work are as follows:

\begin{itemize}
    \item We propose to learn and refine the haze density feature of a hazy image by utilizing the difference information between a generated proposal image ($\textbf{P}$) and a hazy input ($\textbf{I}$) and an end-to-end method named DFR-Net is designed to achieve density-aware dehazing.
    \item We extract haze density features both globally and locally. In GB, a Global Block is introduced, which can explore the image features of $\textbf{P}$ and $\textbf{I}$. To highlight the features that can better describe the density information, a GDFR module is proposed. In LB, local density features are extracted from the dehazing residual between $\textbf{P}$ and $\textbf{I}$. Additionally, a IDRF module is presented to refine local density features stage by stage.
    \item Sufficient experiments are conducted on our DFR-Net and demonstrate that it can achieve better results over the existing state-of-the-art (SOTA) methods on multiple commonly used datasets.
\end{itemize}


\begin{figure}[t]
    \centering
    \includegraphics[width=\linewidth]{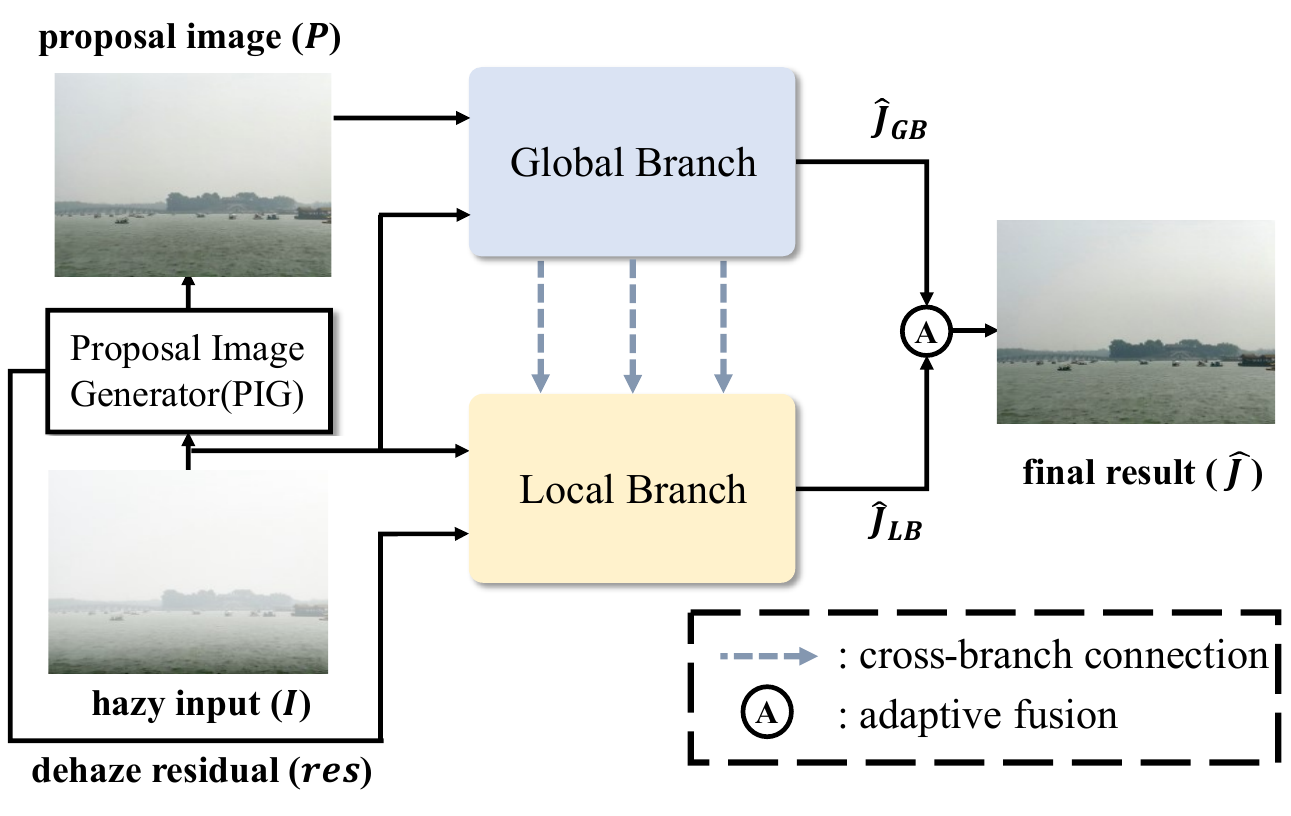}
    \caption{The pipeline of DFR-Net. DFR-Net first generates a proposal image ($\textbf{P}$) by PIG, and $\textbf{P}$ is input to the subsequent two-branch network together with the hazy input ($\textbf{I}$). Each branch predicts a pseudo-clear image and we perform an adaptive fusion to obtain the final result. Note that the global density features in Global Branch are fed into Local Branch by cross-branch connections.}
    \label{fig:pipeline}
    \vspace*{-15pt}
\end{figure}

\section{Related Works}
\subsection{Density-aware Dehazing Methods}

In recent years, several methods \cite{zhang2021hierarchical,deng2020hardgan,guo2022image,yang2022self,wang2021haze,yi2022two,yeperceiving} have attempted to improve the dehazing performance by enabling the network to perceive haze density.

\subsubsection{Density-awareness via estimating T-map} Haze density is influenced by several factors and is inversely proportional to T-map, so some methods learn density information by estimating T-map. Lou et al. \cite{lou2020integrating} predict a T-map first for nighttime image dehazing. Zhang et al. \cite{zhang2021hierarchical} estimate a low-resolution T-map and then jointly input the feature map and the estimated T-map to a Laplacian pyramid decoder to achieve a restored image. Yang et al. \cite{yang2022self} propose a semi-supervised method that does not require paired data. The method estimates T-map, scattering coefficient, and depth to reconstruction hazy images and restores clear images. However, these methods require additional labeled data and might be inaccurate due to the complexity of practical scenes \cite{li2017aod}.

\begin{figure*}[t]
    \centering
    \includegraphics[width=\linewidth]{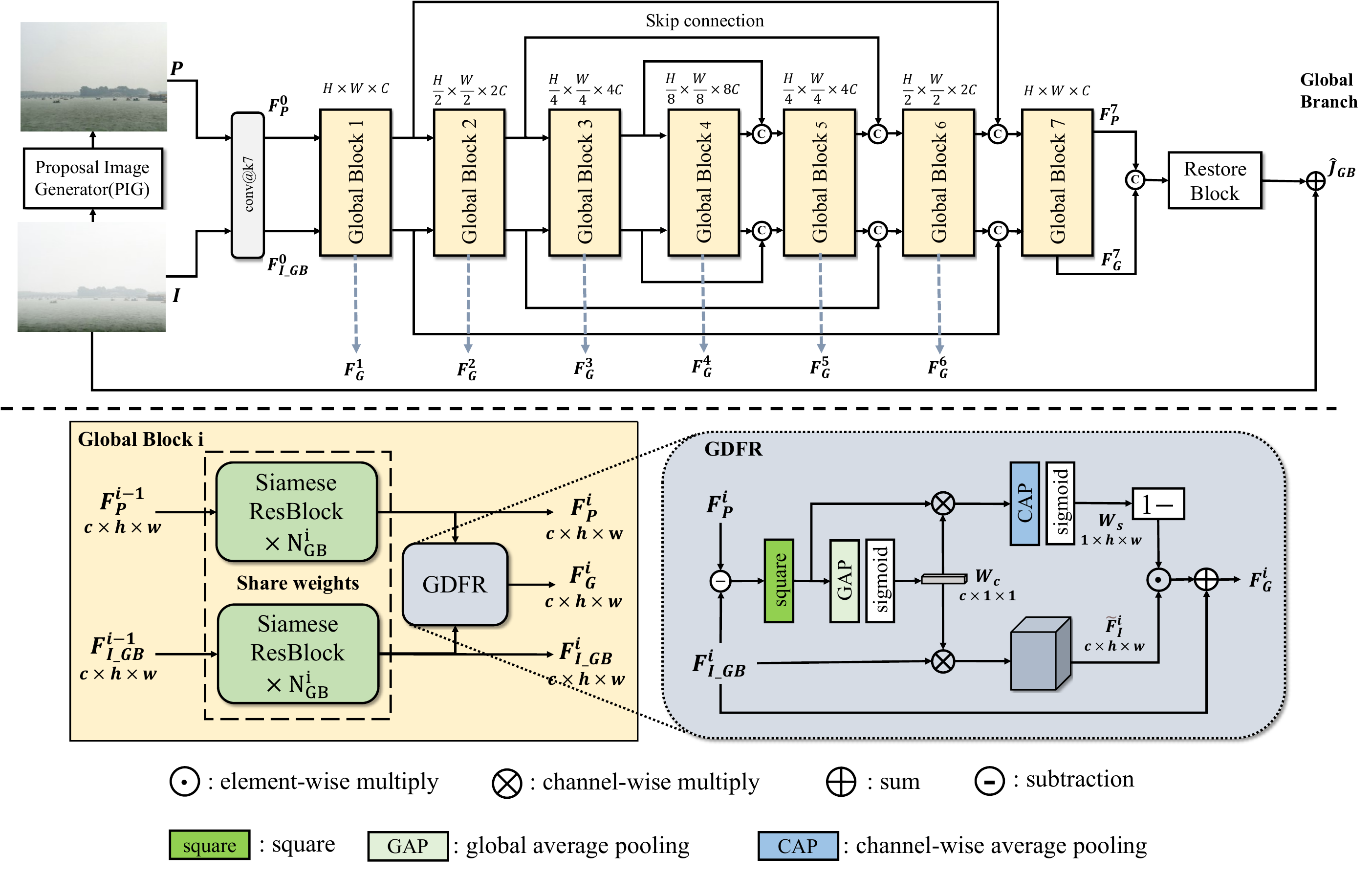}
    \captionsetup{justification=centering}
    \caption{The illustration of GB, global block and GDFR module. Note the global density features are fed into LB by cross-branch connections.}
    \label{fig:GB}
    \vspace*{-10pt}
    \end{figure*}

\subsubsection{Density-awareness via extracting density features directly} Research works \cite{deng2020hardgan,chen2020unsupervised,yeperceiving} directly learn haze density information without estimating a T-map. Deng et al. \cite{deng2020hardgan} design a Haze-Aware Representation Distillation (HARD) module to extract global brightness and a haze-aware map. Chen et al. \cite{chen2020unsupervised} propose an attention mechanism based on dark channel prior to describe haze concentration. However, not estimating the T-map would result in a lack of a comparator to measure density. Generating intermediate results and using the information contained therein can address this issue.

\subsection{Dehazing Methods with Intermediate Results}
Considering the difficulty of recovering images directly from the haze input, dehazing methods \cite{bai2022self,chen2021desmokenet,yeperceiving,hong2022uncertainty} which generate intermediate results (or one result) inside the network to facilitate the dehazing process are proposed. Bai et al. \cite{bai2022self} first generate a reference image by a deep pre-dehazer, and then develop a progressive feature fusion module to fuse the hazy and reference features, which achieves high metrics on several datasets. Chen et al. \cite{chen2021desmokenet} first remove light and thick smoke by a Smoke Remove Network (SRN) to gain a coarse output, which is concatenated with the original input and fed to a Pixel Compensation Network (PCN) to recover the missing pixels in the thick smoke. Hong et al. \cite{hong2022uncertainty} propose an Uncertainty-Driven Dehazing Network. In this method, intermediate results are together generated with uncertainty maps for uncertainty features extraction. Ye et al. \cite{yeperceiving} also pre-generate a pseudo-haze-free image. The hazy input and the pseudo-haze-free image are concatenated to estimate a Density Encoding Matrix describing the relationship between haze density and absolute position and mixed up to the following deep layers.

Despite the above methods extracting feature from intermediate results, they do not fully consider the differences between these results and the haze inputs, especially the differences in haze density. Simple concatenation \cite{chen2021desmokenet,bai2022self} or linear summation \cite{yeperceiving} might lead the networks to rely on the uncertain learning process and lose the capture of information about the differences between the two images. In addition, the lack of a targeted design that addresses the relationship between the intermediate results and the original input leads the extracted features not fine enough and limits the dehazing performance.

Our DFR-Net improves on the aforementioned methods by exploring and refining density features through the utilization of density differences between a generated proposal image and the hazy input, thereby achieving an awareness of haze density and superior dehazing performance.

\section{Method}
As illustrated in Fig. \ref{fig:pipeline}, DFR-Net generates a proposal image using a Proposal Image Generator (PIG), which is a simple U-Net, to facilitate density-awareness. Besides, DFR-Net consists of two primary parts: Global Branch (GB) and Local Branch (LB). The GB and LB are responsible for learning and refining global and local haze density features, respectively, and generate a pseudo result each. In the end, an adaptive fusion is used to obtain the final dehazing result. This section presents a detailed introduction of our method's main ideas.

\begin{figure*}[t]
    \centering
    \includegraphics[width=\linewidth]{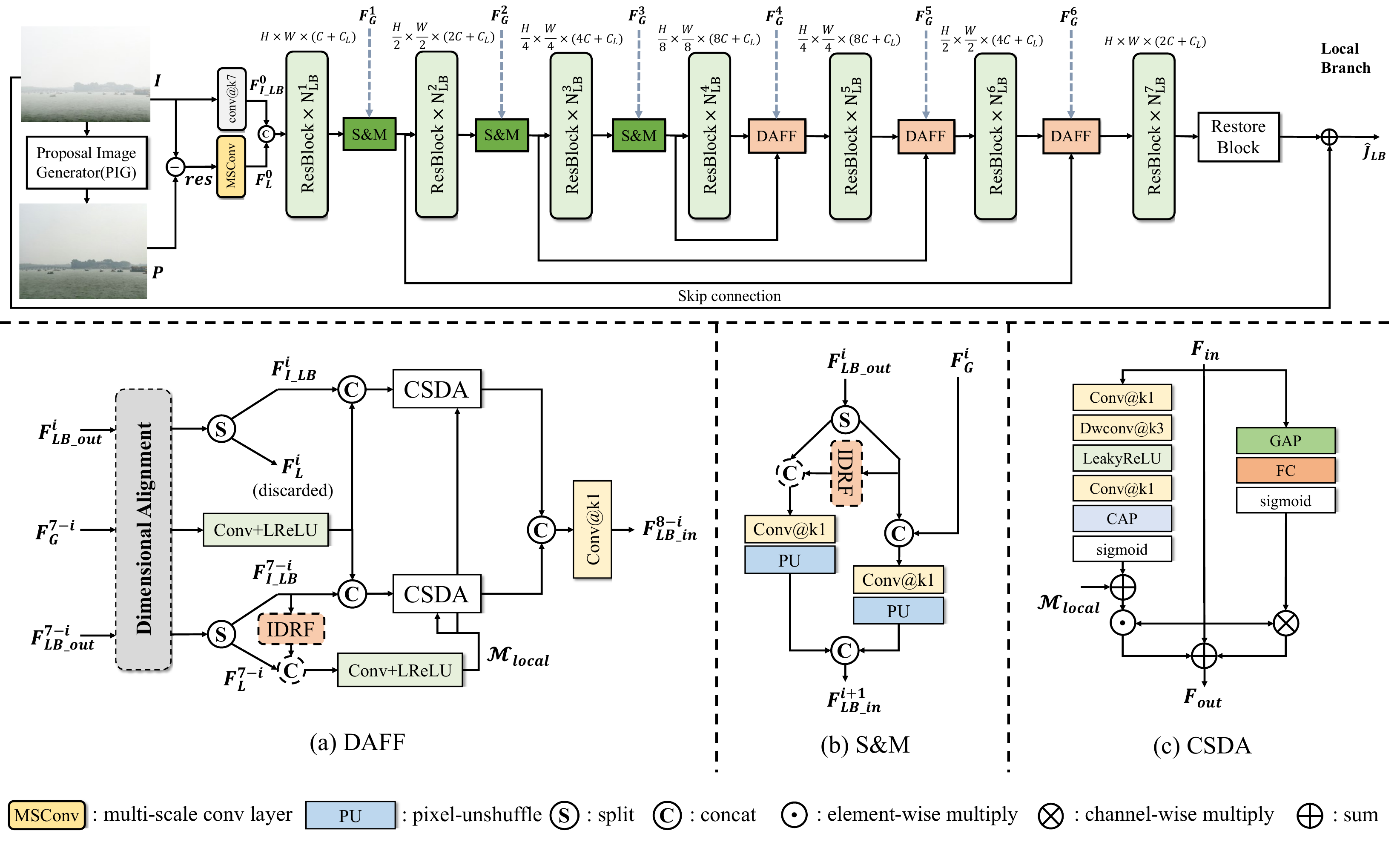}
    \caption{The illustration of the structure of LB (top) and DAFF (a) , S\&M (b), CSDA (c)  modules. The locations of the IDRF usage are indicated by dashed lines representing plug-and-play availability.}
    \label{fig:DAFF}
    \vspace*{-10pt}
  \end{figure*}

\subsection{Proposal Image for Density-awareness}

In DFR-Net, we first generate a proposal image (\textbf{P}) that provides information about the difference in haze density. \textbf{P} exhibits lower overall density than the input image (\textbf{I}), and inconsistent dehazing performance in some local areas. Taking these characteristics into account, we are motivated to extract and refine haze density features in the subsequent network.

To generate \textbf{P}, we employ a Proposal Image Generator (PIG), which is a simple U-net comprising multiple Residual Blocks (ResBlocks). We pre-train PIG with paired images and incorporate it into subsequent branches as an end-to-end network.

\subsection{Global Branch}

\subsubsection{Overview} The Global Branch (GB) is designed to extract and refine global density features using the overall haze density difference. As depicted in Fig. \ref{fig:GB} (top), GB consists of a 7-stage U-Net with 7 global blocks, where each block has a Siamese structure with $N_{GB}^{i}(i\in[1,...,7])$ ResBlocks for feature extraction and a GDFR module for global density feature refinement. The inputs to GB are \textbf{I} and \textbf{P} $\in \mathbb{R}^{H\times W\times 3}$, which are embedded to $F^{0}_{I\_GB}$, $F^{0}_P$ $\in \mathbb{R}^{H\times W\times C}$ using a convolutional layer. Upsampling or downsampling is performed between every two blocks. To obtain the predicted dehazing residual, the output features of the last global block are fed to a restore block, and the predicted dehazing residual is added to \textbf{I} to obtain a pseudo result, $\hat{J}_{GB}$.

We design the global block as a basic unit in GB, taking the haze density relationship between \textbf{I} and \textbf{P} into consideration. Specifically, the $i$-th global block is composed of a Siamese structure with $N_{GB}^{i}$ ResBlocks for feature extraction and a GDFR module for global density feature refinement. For the $i$-th global block, the inputs are the outputs of the previous block, $F^{i-1}_{I\_GB}$ and $F^{i-1}_P$, the outputs are $F^{i}_{I\_GB}$, $F^{i}_P$ and the refined global density feature, $F^{i}_G$.

\subsubsection{Siamese ResBlocks for global feature extraction} Siamese structure shares weights and can measure similarity \cite{bertinetto2016fully,he2018twofold,guo2017learning} or dissimilarity \cite{koch2015siamese,wu2017face} between samples effectively. So we utilize a Siamese structure to extract image features from both \textbf{P} and \textbf{I}, which enables the network to establish a relationship between images with different haze densities. Moreover, compared to a single hazy input structure, the Siamese structure enables the features extracted by the network to better perceive the variation in haze density.

\subsubsection{GDFR for global feature refinement} To further refine the extracted features, we propose the Global Density Feature Refinement (GDFR) module, which aims to highlight the features that better describe global density information and thus pull apart the features of images with different densities. For the implementation, we compute the difference between the feature maps of \textbf{I} and \textbf{P}, and then square each element in the difference map. This subtractive operation filters out non-density information, and the resulting feature channels with larger differences can better describe density information. We perform global average pooling (GAP) and sigmoid operation on the feature differences to obtain density-related channel weights, denoted as $W_c$. We then multiply these weights with $F^{i}_{I\_GB}$ channel-wise: 
\begin{equation}
  \Tilde{F}_I^i=F_{I\_GB}^i\otimes\sigma(GAP(POW((F_P^i-F_{I\_GB}^i),2))) 
\end{equation}
where $GAP(\cdot)$ stands for global average pooling, $\sigma(\cdot)$ represents sigmoid operation, $\otimes$ stands for channel-wise multiply and $POW(a,b)$ represents a power of exponent $b$ for each element of $a$.

In spatial dimension, we first multiply the channel weights by the squared feature difference. Next, we apply channel-wise average pooling (CAP) and sigmoid operations to obtain a 2-D weight map that represents the spatial density difference $W_s$. Higher value in $W_s$ indicates that haze has been more effectively removed here. Therefore, $1-W_s$ can represent an attention map that guides the network's focus. Here $1$ denots a 2-D tensor of ones with the same shape of $W_s$. We then multiply $1-W_s$ and $\Tilde{F}_I^i$ and add the result to $F^{i}_{I\_GB}$ to obtain the finally refined global density feature ${F}_G^i$:

\begin{equation}
  F_G^i=(1-\sigma(CAP(POW((F_P^i-F_{I\_GB}^i),2))))\odot\Tilde{F}_I^i+F^{i}_{I\_GB}
\end{equation}
where $CAP(\cdot)$ stands for channel-wise average pooling.

After the last block, we concatenate $F^{7}_P$ and $F^{7}_G$ and input the concatenated feature to a restore block, which is composed of four ResBlocks and a convolutional layer, to obtain the dehazing residual.
\begin{equation}
  res_{GB}=RB_{GB}(cat(F_P^{7},F_G^{7})) 
\end{equation}
where $RB_{GB}(\cdot)$ denotes the calculation of the restore block in GB and $cat(\cdot)$ denotes the channel-wise concatenation operation. And the pseudo-clear image of GB can be obtained by: $\hat{J}_{GB} = I + res_{GB}$.

\subsection{Local Branch}

\subsubsection{Overview} LB extracts local density features from the dehazing residual of \textbf{P} and \textbf{I} , which contains local density differences and indicates hard dehazing or high haze density areas. To refine these local features, we propose the IDRF module that gradually adjusts the features to match those of clear images.

The main body of LB is a 7-stage U-net, as illustrated in Fig. \ref{fig:DAFF} (top). Each stage is composed of $N_{LB}^i (i\in[1,...,7])$ ResBlocks. LB takes \textbf{I} and \textbf{res}$\in \mathbb{R}^{H\times W\times 3}$ as inputs and embeds them to shallow features $F^{0}_{I\_LB} \in \mathbb{R}^{H\times W\times C}$, $F^{0}_L\in \mathbb{R}^{H\times W\times C_L}$ using a normal convolutional layer and a multi-scale convolutional layer, respectively. In the encoder stages, local and image features are extracted by a split and merge way, and in the decoder stages, DAFF fuses shallow image features, deep image features, local features, and refined global features from GB. The dehazing residual can be obtained by a restore block, and the pseudo result $\hat{J}_{LB}$ can be obtained by summing it and \textbf{I}.

\subsubsection{S\&M for local feature extraction} The local features are extracted in an S\&M way. Firstly, $F^{0}_{I\_LB}$ and $F^{0}_L$ are merged by concatenation and fed into stage-1, resulting in an output feature with ($C+C_L$) channels. The output feature is then processed by the S\&M module. As shown in Fig. \ref{fig:DAFF} (b), the input feature is split into an image feature and a local density feature. Notably, the number of channels for the image feature varies from stage to stage, while the number of channels for the local feature is consistently fixed at $C_L$. To utilize the global density information obtained from GB, the image feature is concatenated with the refined global feature from GB. Afterwards, the concatenated feature and local feature are downsampled by $1\times 1$ convolutional layers and pixel-unshuffle, respectively. The downscaled features are then concatenated and fed into the next stage. This design allows local features to fully interact with hazy image features, thus enhancing the reliability of local features.

\subsubsection{DAFF for feature fusion}\label{sec:method} Several methods introduce skip connection to aggregate shallow and deep features and simply concatenate or add them together \cite{zamir2022restormer,yin2021attentive}, which might result in a loss of information. To fully utilize the information relevant to density, we employ Density Aware Feature Fusion (DAFF). Specifically, given a set of features: $\{F^i_{LB\_out}, F^{7-i}_{LB\_out}, F^{7-i}_G\} (i\in[1, 2, 3])$, which represent the output feature of the $i$-th, $(7-i)$-th stage of LB and the refined global feature of the $(7-i)$-th block in GB, we first align them to a same shape and split the image features and local density features. Then we perform convolutions and LeakyReLU (Conv+LReLu) on $F^{7-i}_G$ and the results are concatenated with $F^i_{I\_LB}$ and $F^{7-i}_{I\_LB}$ respectively and input to Channel-Spatial Density Attention (CSDA) modules, as illustrated in Fig. \ref{fig:DAFF} (c). Moreover, the local feature $F^{7-i}_L$ is projected to a 2-D local attention map ($\mathcal{M}_{local}$) by Conv+LReLu and $\mathcal{M}_{local}$ is fed to CSDAs. Finally, the outputs of CSDAs are concatenated and compressed to the channel number of the subsequent stage by a $1\times 1$ convolutional layer. 

\begin{figure}[t]
  \centering
  \includegraphics[width=\linewidth]{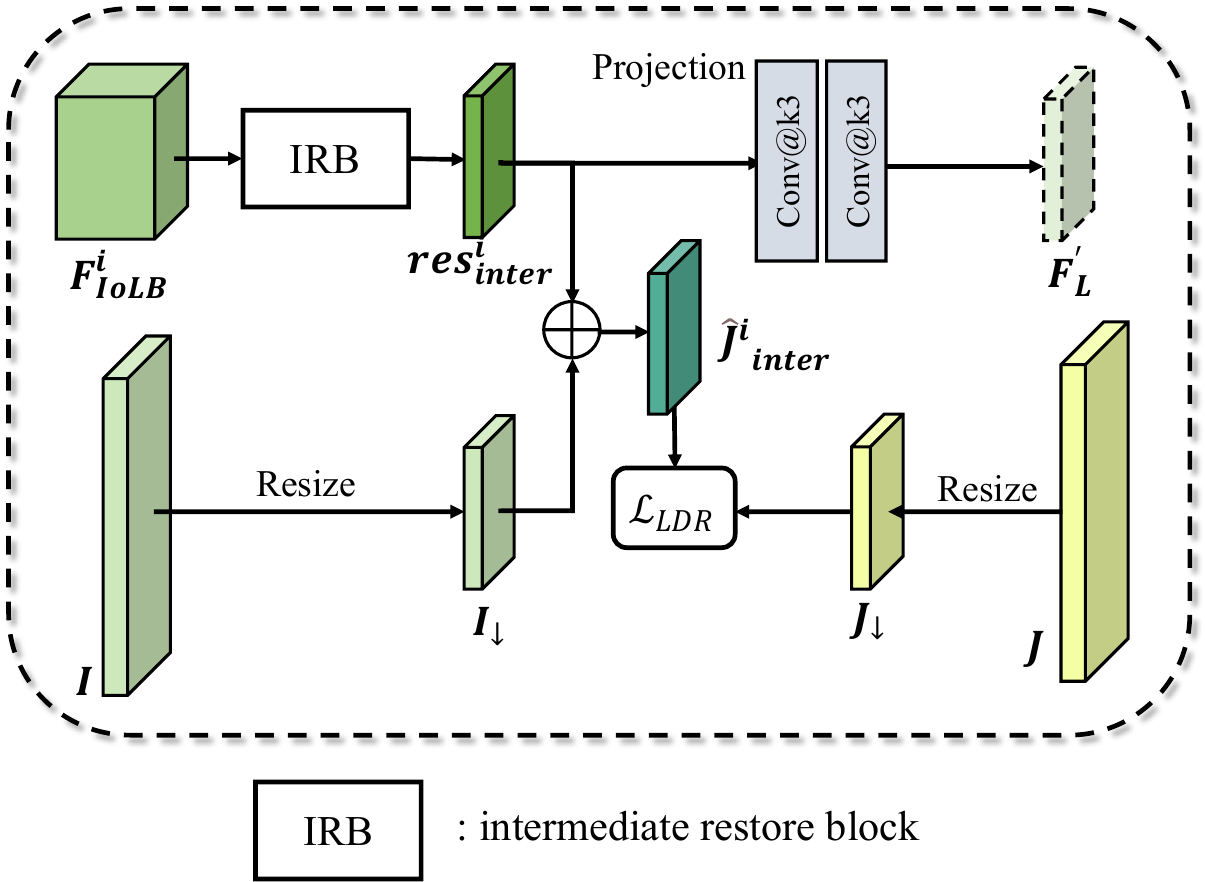}
  \caption{The illustration of IDRF module. The projected feature $F^{'}_L$ will be used to update local features in corresponding S\&M or DAFF module.}
  \label{fig:IDRF}
  \vspace*{-5pt}
\end{figure}

\begin{table*}[ht]
    \caption{Quantitative comparison of DFR-Net with the state-of-the-art image dehazing methods on different datasets (PSNR (DB)/SSIM). Best results are \textbf{bolded} and second best results are \underline{underlined}. Cells where results are not available are replaced by "-"}
    \begin{center}
    \begin{tabular}{c c c c c c c c c c}
    \toprule[1pt]
    \multicolumn{2}{c}{\multirow{2}{*}{Method}} & \multicolumn{2}{c}{RESIDE-outdoor} & \multicolumn{2}{c}{Haze4K} & \multicolumn{2}{c}{NH-HAZE} & \multicolumn{2}{c}{Dense-Haze}\\ \cmidrule(l){3-4} \cmidrule(l){5-6} \cmidrule(l){7-8} \cmidrule(l){9-10}
    \multicolumn{2}{c}{} & PSNR(dB) & SSIM & PSNR(dB) & SSIM & PSNR(dB) & SSIM & PSNR(dB) & SSIM\\
    \midrule
    \multirow{13}{*}{Non-Density-Aware} & DCP \cite{he2010single} (TPAMI' 10) & 19.13 &0.815 & 14.01 & 0.760 & 10.57 & 0.522 & 11.01 & 0.416\\
    & DehazeNet \cite{cai2016dehazenet} (TIP' 16) & 24.75 & 0.927 & 19.12 & 0.840 & 12.86 & 0.545 & 9.48 & 0.438\\
    & AOD-Net \cite{li2017aod} (ICCV' 17) & 24.14 & 0.920 & 17.15 & 0.830 & 15.40 & 0.571 & 12.82 & 0.468\\
    & GDNet \cite{liu2019griddehazenet} (ICCV' 19) & 30.86 & 0.982 & 23.29 & 0.930 & 18.33 & 0.667 & 14.96 & 0.530\\
    & MSBDN \cite{dong2020multi} (CVPR' 20) & 33.48 & 0.982 & 22.99 & 0.850 & 19.23 & 0.713 & 15.13 & 0.555\\
    & FFA-Net \cite{qin2020ffa} (AAAI' 20) & 33.57 & 0.984 & 26.96 & 0.950 & 19.87 & 0.694 & 12.22 & 0.444\\
    & AECR-Net \cite{wu2021contrastive} (CVPR' 21) & - & - & - & - & 19.88 &0.722 & 15.80 & 0.466 \\ 
    & CEEF \cite{liu2021joint} (TMM' 22) & 19.13 & 0.792 & -& -& -& - &- &-\\
    & SGID-PFF \cite{bai2022self} (TIP' 22) & 30.20 & 0.975 &  - & - &  - & - & - & -\\ 
    & UDN \cite{hong2022uncertainty} (AAAI' 22) & 34.92 & 0.987 & - & - &  - & - & - & - \\ 
    & QCNN-H \cite{frants2023qcnn} (TC' 23) & 28.74 & 0.964 & - & - & - & - & - & - \\
    & MFINEA \cite{sun2023multi} (NN' 23) & 33.88 & 0.981 & - & - & - & - & \underline{18.34} & \underline{0.609} \\
    & DehazeFormer-B \cite{song2023vision} (TIP' 23) & 34.95 & 0.984 & 30.29 & \underline{0.985} & 17.37 & 0.725 & - & - \\ \midrule

    \multirow{4}{*}{Density-Aware} & HDDNet \cite{zhang2021hierarchical} (TC' 22) & 22.52 & 0.910 & - & -&  - & - & - & -\\
    & DeHamer \cite{guo2022image} (CVPR' 22) & \underline{35.18} & 0.986 & - & -& \underline{20.66}&0.684 & 16.62 & 0.560 \\
    & PMNet \cite{yeperceiving} (ECCV' 22) & 34.74 & \underline{0.990} & \underline{33.49} & 0.980 & 20.42 & \underline{0.731} & 16.79 & 0.510 \\
    \cmidrule{2-10}
    & DFR-Net (ours) & \textbf{35.34} & \textbf{0.993} & \textbf{34.63} & \textbf{0.993} & \textbf{21.21} & \textbf{0.810} & \textbf{18.85} & \textbf{0.674}\\
    \bottomrule
    \end{tabular}
    \end{center}
    \label{tab:result}
    \end{table*}
    
\begin{figure*}[t]
\centering
\includegraphics[width=\linewidth]{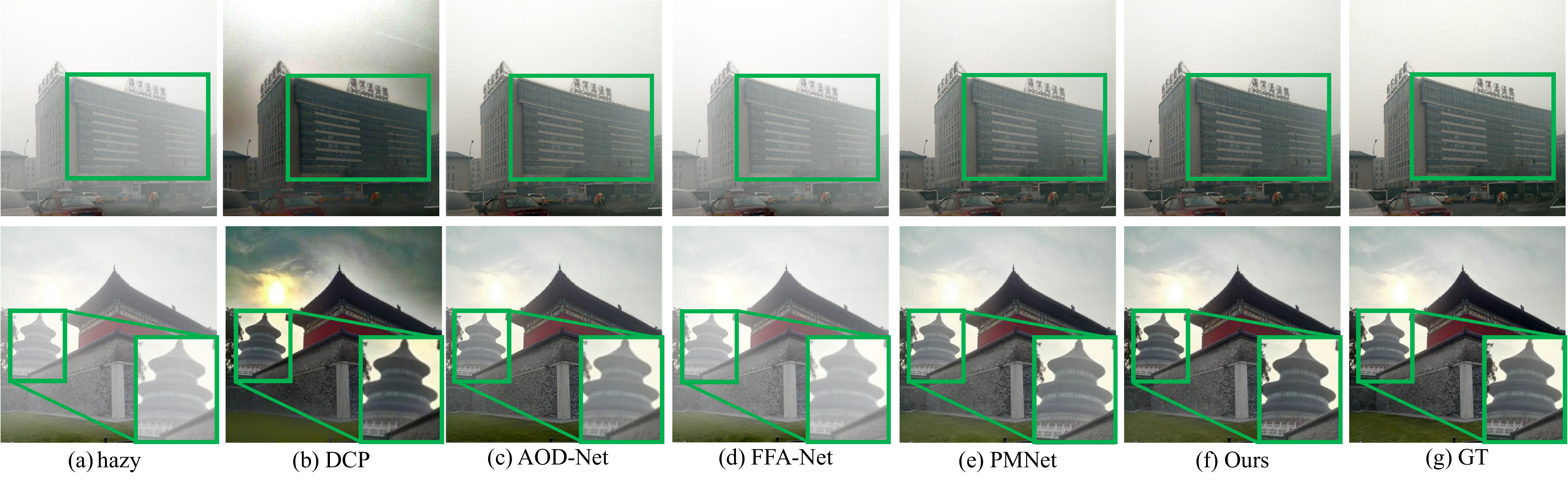}
\centering
\caption{Visual comparison of various methods on Haze4K \cite{liu2021synthetic} dataset. Areas where our method works better are boxed out and zoomed in, or you can zoom in yourself to get a better view.}
\label{fig:h4k_compare}
\end{figure*}


\begin{figure*}[ht]
\centering
\includegraphics[width=\linewidth]{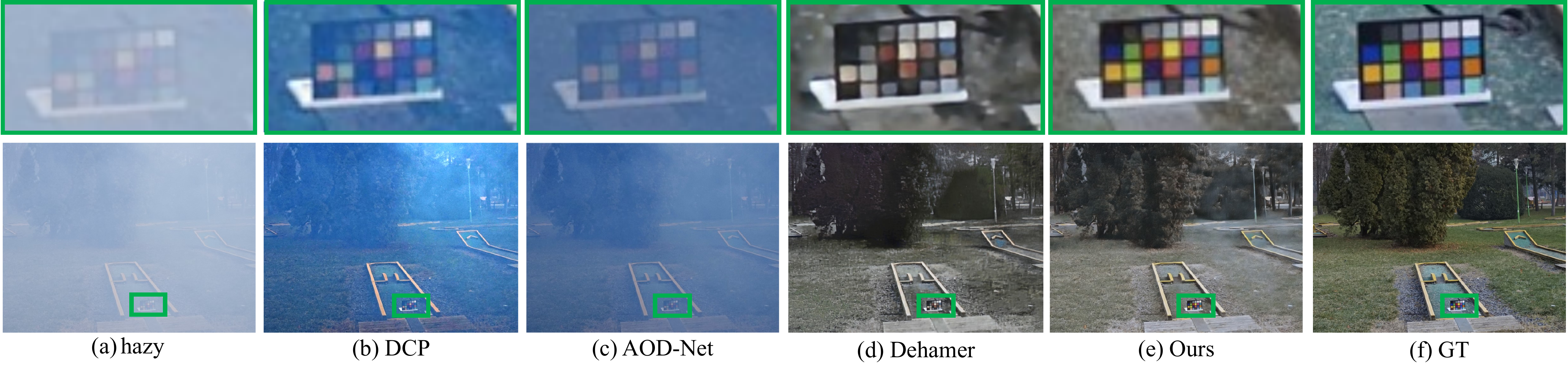}
\centering
\caption{Visual comparison of various methods on Dense-Haze \cite{ancuti2019dense} dataset. Areas where our method works better are boxed out and zoomed in, or you can zoom in yourself to get a better view.}
\label{fig:Dense_compare}
\end{figure*}

\subsubsection{IDRF for local feature refinement} To refine the local density feature, we further introduce IDRF module. As shown in Fig. \ref{fig:IDRF}, IDRF takes $F^i_{I\_LB}$ as input to obtain intermediate dehazing residual $res_{inter}^i$ by an intermediate restore block (IRB) composed of two ResBlocks and a convolutional layer. Then the $res_{inter}^i$ is projected to a $C_L$-channel feature embedding $F_L^{'}$ which is subsequently concatenated with local features as shown in Fig. \ref{fig:DAFF} (a) and (b). With this process, local density features can be updated by the current dehazing residual. To optimize this module and pull local features in towards clear image features, we employ a local density refinement loss ($\mathcal{L}_{LDR}$) which will be introduced in Sec. \ref{section:loss}.

Similar to GB, LB predicts a pseudo-clear result: $\hat{J}_{LB} = I + res_{LB}$.
Finally we fuse the pseudo-results of the two branches with a learnable parameter $\alpha $: $\hat{J} = \alpha\times\hat{J}_{GB}+(1-\alpha)\times\hat{J}_{LB}$.

\subsection{Loss Function} \label{section:loss}
The overall loss fuction of our DFR-Net can be formulated as: $\mathcal{L}=\mathcal{L}_{Rec}+\lambda_1\mathcal{L}_P+\lambda_2\mathcal{L}_{RD}+\lambda_3\mathcal{L}_{LDR}$, where $\mathcal{L}_{Rec}$ and $\mathcal{L}_P$ denote L1 loss and perceptual loss \cite{DBLP:journals/corr/JohnsonAL16} between the predicted haze-free output $\hat{J}$ and ground truth $J$, $\mathcal{L}_{RD}$ and $\mathcal{L}_{LDR}$ represent representation dissimilarity loss and local density refinement loss, and $\lambda_1$, $\lambda_2$ and $\lambda_3$ are hyper-parameters for loss regulation.

\subsubsection{Representation Dissimilarity Loss}
In this paper, we introduce a Siamese structure to learn density-related features from \textbf{I} and \textbf{P}. To motivate the Siamese structure to learn more information about the differences between the inputs, we design representation dissimilarity loss:
\begin{equation}
    \mathcal{L}_{RD}=\sum_{i=1}^{n}\langle F_P^i, F_{I\_GB}^i\rangle 
\end{equation}
where $\langle a, b\rangle $ represents the calculation of the cosine similarity between $a$ and $b$. We compute cosine similarities between the intermediate features of \textbf{I} and \textbf{P} and minimize them to let the Siamese structure learn more representation about the difference between the two inputs.

\subsubsection{Local Density Refinement Loss}
To achieve the pulling of local features from haze images to clear images, we introduce local density refinement loss.
\begin{equation}
    \mathcal{L}_{LDR}=\frac{1}{k}\sum_{j=1}^{k}\Vert \hat{J}^j_{inter}(x)-Down^j(J(x))\Vert_1
\end{equation}
where $k$ denotes the number of IDRF modules which are applied, $\hat{J}_{inter}^j(x)$ is the $j$-th intermediate predicted clear output, and $Down^j(\cdot)$ is the operation that downsamples ground truth to the size of the corresponding intermediate output.

\begin{figure*}[t]
    \centering
    \includegraphics[width=\linewidth]{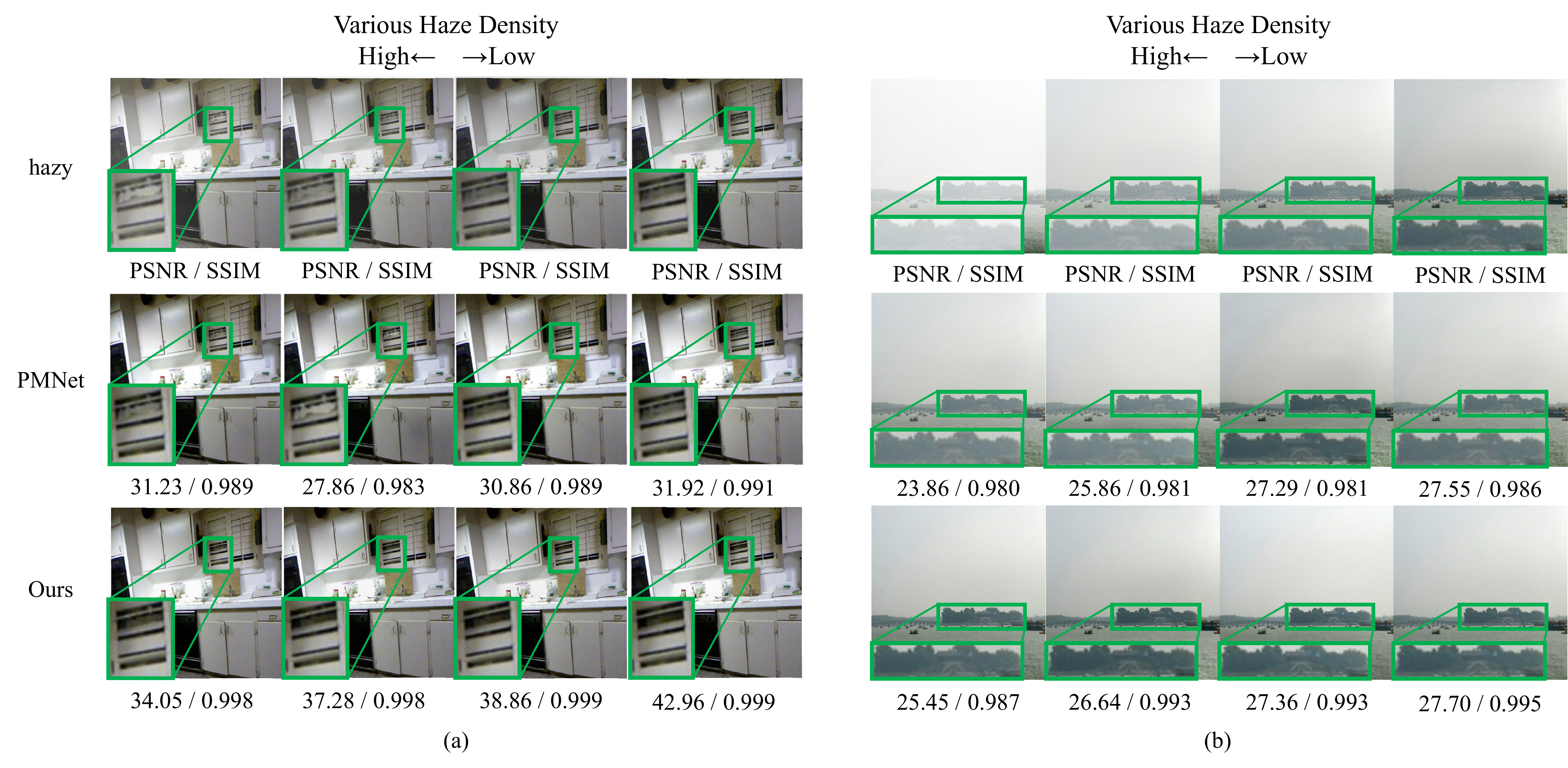}
    \caption{Visual comparison with PMNet on Haze4K \cite{liu2021synthetic} dataset. Note that the densities of the input images are varied for each set of images of the same scene, gradually decreasing from left to right. Please zoom in for a better view.}
    \label{fig:vis_h4k}
  \end{figure*}


\section{Experiments}

\subsection{Datasets and Metrics}
We conduct experiments on several datasets to train our method and test our method's dehazing performance. The datasets include: (1) RESIDE outdoor \cite{li2018benchmarking}, which contains 313950 synthetic outdoor hazy/clear image pairs for training and 500 pairs for testing; (2) Haze4K \cite{liu2021synthetic}, which includes a training set of 3000 indoor-outdoor mixed image pairs and a testing set of 1000 image pairs; (3) NH-HAZE \cite{lugmayr2020ntire}, a real-world dataset for the NTIRE 2020 competition, which consists of 55 pairs of non-homogeneous hazy images and clear images of real scenes (45, 5 and 5 pairs for training, validation and test respectively); (4) Dense-Haze \cite{Ancuti_2019_CVPR_Workshops, ancuti2019dense}, a real-world dataset for the NTIRE 2019 competition and contains 55 pairs of dense-haze images and corresponding clear images (with same data split as NH-HAZE). We evaluate the dehazing effectiveness of our method using two commonly used image quality metrics: PSNR (dB) and SSIM.

\subsection{Implementation Details}
\subsubsection{Network Configuration}
In our work, we use the Basic Block structure proposed in \cite{qin2020ffa} as our ResBlock. $\{N_{GB}^1,$ $N_{GB}^2,$ $N_{GB}^3,$ $N_{GB}^4,$ $N_{GB}^5,$ $N_{GB}^6,$ $N_{GB}^7\}$ and $\{N_{LB}^1,$ $N_{LB}^2,$ $N_{LB}^3,$ $N_{LB}^4,$ $N_{LB}^5,$ $N_{LB}^6,$ $N_{LB}^7\}$ are set to $\{2,$ $2,$ $3,$ $4,$ $3,$ $2,$ $2\}$ and $\{4,$ $6,$ $8,$ $10,$ $6,$ $8,$ $8\}$. In particular, we set the basic feature dimension of GB and LB, $C$, to 32 and the local density dimension $C_L$ to 4. All upsampling or downsampling operations are implemented by $1\times1$ convolution with pixel-shuffle or pixel-unshuffle. And we use IDRF module after every stage in LB except the last stage. 

\subsubsection{Training Settings}
The PIG is pre-trained and spliced with GB and LB. And the whole network is trained in an end-to-end fashion. We use AdamW optimizer ($\beta_1=0.9$, $\beta_2=0.999$, weight decay is $1e^{-4}$) to train the model and iterate 600k times with the initial learning rate $1e^{-4}$ reduced to $1e^{-6}$ with the cosine annealing \cite{loshchilov2016sgdr}.
Following \cite{zamir2022restormer}, we perform progressive learning in our training process, which leads the network to adapt to those inputs close to the size in practical applications. For loss regulation, we set $\lambda_1=0.2$, $\lambda_2=0.001$ and $\lambda_3=0.1$. The data augmentations include random cropping, horizontal flipping, and vertical flipping.

\subsection{Comparison with State-of-the-art Methods}
Our comparison methods include DCP \cite{he2010single}, DehazeNet \cite{cai2016dehazenet}, AOD-Net \cite{li2017aod}, GDNet \cite{liu2019griddehazenet}, MSBDN \cite{dong2020multi}, FFA-Net \cite{qin2020ffa}, AECR-Net \cite{wu2021contrastive}, CEEF \cite{liu2021joint}, SGID-PFF \cite{bai2022self}, UDN \cite{hong2022uncertainty}, QCNN-H \cite{frants2023qcnn}, MFINEA \cite{sun2023multi} , DehazeFormer \cite{song2023vision}, HDDNet \cite{zhang2021hierarchical}, DeHamer\cite{guo2022image} and PMNet \cite{yeperceiving}. 

\subsubsection{Quantitative Evaluations} \label{Quantitative Evaluations}

The quantitative results are shown in Table \ref{tab:result}. It demonstrates that our method achieves the highest metrics on RESIDE-outdoor, Haze4K, Dense-Haze and NH-HAZE datasets, outperforming other SOTA methods. Notably, DFR-Net achieves a 1.14dB PSNR gain on the Haze4K dataset and significant improvements in SSIM on all datasets. 

Among the density-aware methods, our DFR-Net outperforms all other methods \cite{deng2020hardgan,zhang2021hierarchical,guo2022image,yeperceiving}. Compared to the transmission-aware (density-related) methods DeHamer \cite{guo2022image} and HDDNet \cite{zhang2021hierarchical}, we directly extract density features from hazy images and obtain better performance. DFR-Net also surpasses PMNet on metrics on multiple datasets by utilizing density difference information between \textbf{I} and \textbf{P}, rather than simply concatenating them.

\subsubsection{Qualitive Evaluations} \label{Qualitive Evaluations}
Fig. \ref{fig:h4k_compare} shows visual comparisons between our DFR-Net and SOTA methods on Haze4K dataset, demonstrating that our DFR-Net can more effectively remove haze than other methods. Specifically, AOD-Net \cite{li2017aod} and FFA-Net \cite{qin2020ffa} still leave haze residue in most areas, DCP \cite{he2010single} removes some of the haze but suffers from color distorion. While PMNet \cite{yeperceiving} achieves better dehazing performance than previous methods, our method is able to recover clearer images.

In addition, we compare the visual quality performance of our method with other methods on real-world dataset (Dense-Haze) in Fig. \ref{fig:Dense_compare}. Compared to other methods, our DFR-Net removes the overall haze while achieving fine restoration in image details. This is attributed to the pulling in of the image features towards a clear image at each stage in LB, which reduces the occurrence of color distortion and blurring of details in our results.

We further compare the visual results of our DFR-Net with those of PMNet \cite{yeperceiving} on the Haze4K dataset. Fig. \ref{fig:vis_h4k} illustrates the dehazing outcomes of PMNet and DFR-Net for both indoor and outdoor images with varying haze densities. DFR-Net exhibits more consistent dehazing performance across different haze densities compared to PMNet. In particular, DFR-Net demonstrates superior performance in preserving local image details, as depicted in Fig. \ref{fig:vis_h4k} (a). Conversely, PMNet struggles to accurately restore local details in the dehazed images. Additionally, DFR-Net outperforms PMNet in handling images with high global haze density, as shown in Fig. \ref{fig:vis_h4k} (b). The incorporation of both global and local density difference information in DFR-Net contributes to its robustness in perceiving and effectively addressing different haze densities. The global density component enables the network to better comprehend variations in haze densities, allowing for improved performance on images with diverse densities. On the other hand, the local density component motivates the network to identify and restore details in regions with high density, resulting in finer dehazed images with enhanced visual quality. Further details and discussions regarding these aspects are presented in Sec. \ref{section:ablation}.

\subsection{Ablation Studies} \label{section:ablation}

The main innovation of DFR-Net is the extraction and utilization of global and local haze density information. Therefore, we conduct ablation experiments and analyses in this section to demonstrate the effectiveness of our utilization of haze density information.

We conduct ablation experiments on the modules and loss functions proposed in this paper  to evaluate their effectiveness on dehazing. We first establish a \textbf{base} network (\ding{172}), which consists of a non-weight sharing GB (pseudo-Siamese structure), and an LB that takes the directly concatenated \textbf{I} and \textbf{P} as input and aggregates features by concatenation. Only $\mathcal{L}_{Rec}$ and $\mathcal{L}_{P}$ are employed to optimize the \textbf{base} network. Then we define several variants to verify the effectiveness of our proposed modules and loss functions on dehazing: \ding{173} \textbf{+Siamese}: Use the Siamese structure for feature extraction. \ding{174} \textbf{+$\mathcal{L}_{RD}$}: Incorporate $\mathcal{L}_{RD}$ into the total loss function. \ding{175} \textbf{+GDFR}: Use GDFR to obtain refined global density features. \ding{176} \textbf{+DAFF}: Aggregate features by DAFF. \ding{177} \textbf{+DR}: Explore local density from \textbf{D}ehazing \textbf{R}esidual (DR) in LB. \ding{178} \textbf{+IDRF}: Use IDRF after each stage of LB, except the last one. \ding{179} \textbf{+$\mathcal{L}_{LDR}$} (our default setting): Include $\mathcal{L}_{LDR}$ in the total loss function. Among them, variants \ding{173}-\ding{175} and \ding{176}-\ding{179} gradually introduce global and local haze density respectively. The ablation results are presented in Table \ref{tab:ablation}. Subsequently, we will analyze the effectiveness of introducing global and local haze density separately.

\captionsetup[table]{font={normalfont}}
\begin{table*}[t]
  \caption{Ablation studies on proposed modules and loss functions on the Haze4K dataset. Note that FLOPs and Params are measured on 256 $\times$ 256 images.}
  \begin{center}
  \begin{tabular}{cccccc}
  \toprule
  &Label&Setting  & PSNR (dB)   & FLOPs (G) & Params (M)\\ \midrule
  w/o density difference & \ding{172}&base &  30.52 &222.46& 35.76 \\ \midrule
  \multirow{3}{*}{+ global density difference}&\ding{173}&+Siamese &  31.21 &241.94& 35.15 \\
  &\ding{174}&+$\mathcal{L}_{RD}$ & 31.55 &241.94& 35.15 \\
  &\ding{175}&+GDFR    &  32.65 &242.63& 35.15 \\ \midrule
  \multirow{4}{*}{+ local density difference}&\ding{176}&+DAFF      & 32.89 &271.54& 40.52 \\
  &\ding{177}&+DR    & 33.01 & 271.56 &40.52\\
  &\ding{178}&+IDRF    & 33.52  & 286.60 &42.11\\
  &\ding{179}&+$\mathcal{L}_{LDR}$ (default) & \textbf{34.63}&  286.60 &42.11\\ \bottomrule
  \end{tabular}

\end{center}

\label{tab:ablation}
\end{table*}

\begin{figure*}[t]
    \centering
    \includegraphics[width=\linewidth]{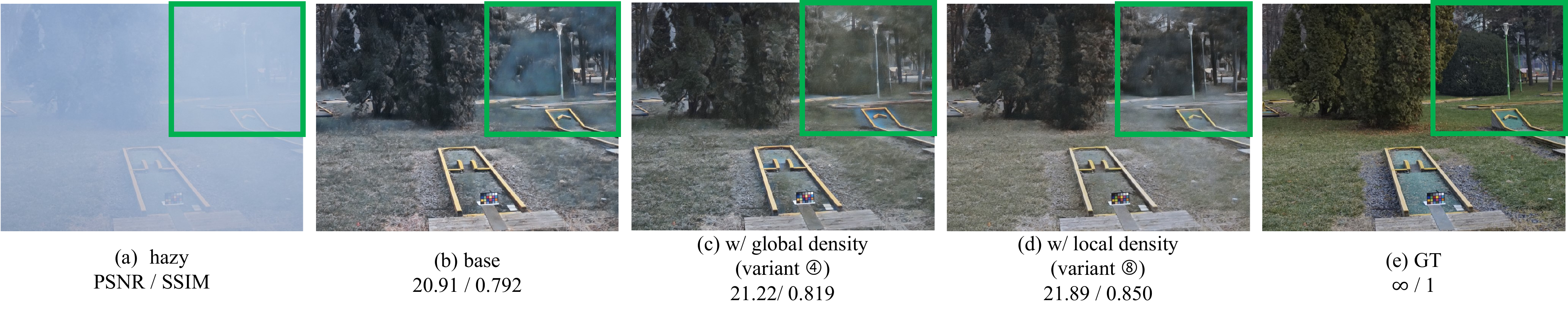}
    \caption{Visualization of the ablation experiments on exploring and utilizing global / local haze density difference. Areas with large variance in dehazing effectiveness are framed out. Please zoom in for a better view.}
    \label{fig:ablation_dense}
\end{figure*}
    
\begin{figure}[t]
    \centering
    \includegraphics[width=\linewidth]{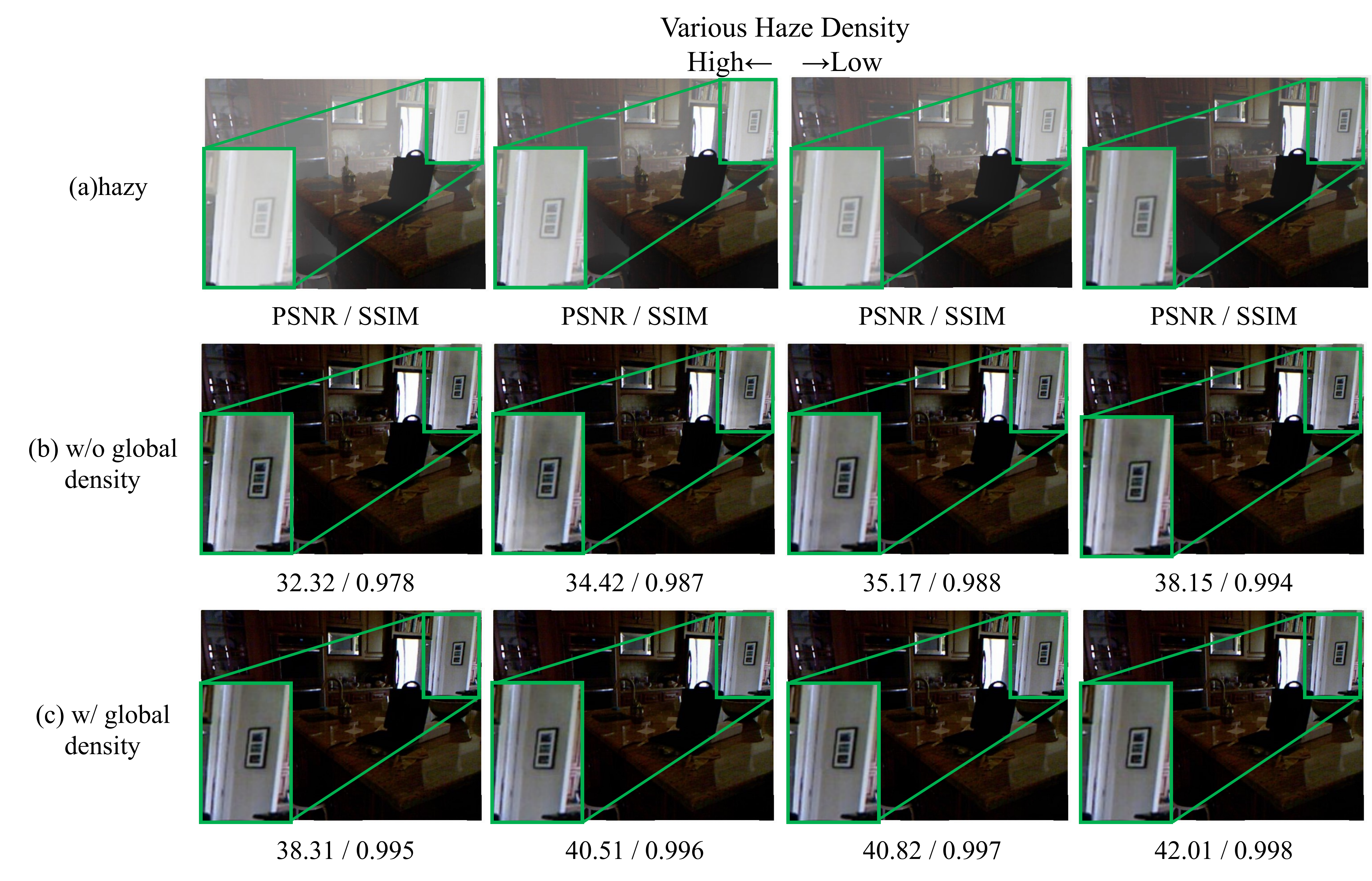}
    \caption{Visualization of the ablation experiments on utilizing global density difference and global feature refinement. Note that the haze densities of the input image are varied and gradually decrease from left to right. Please zoom in for a better view.}
    \label{fig:ablation_global}
    
  \end{figure}

\begin{figure}[t]
  \centering
  \includegraphics[width=\linewidth]{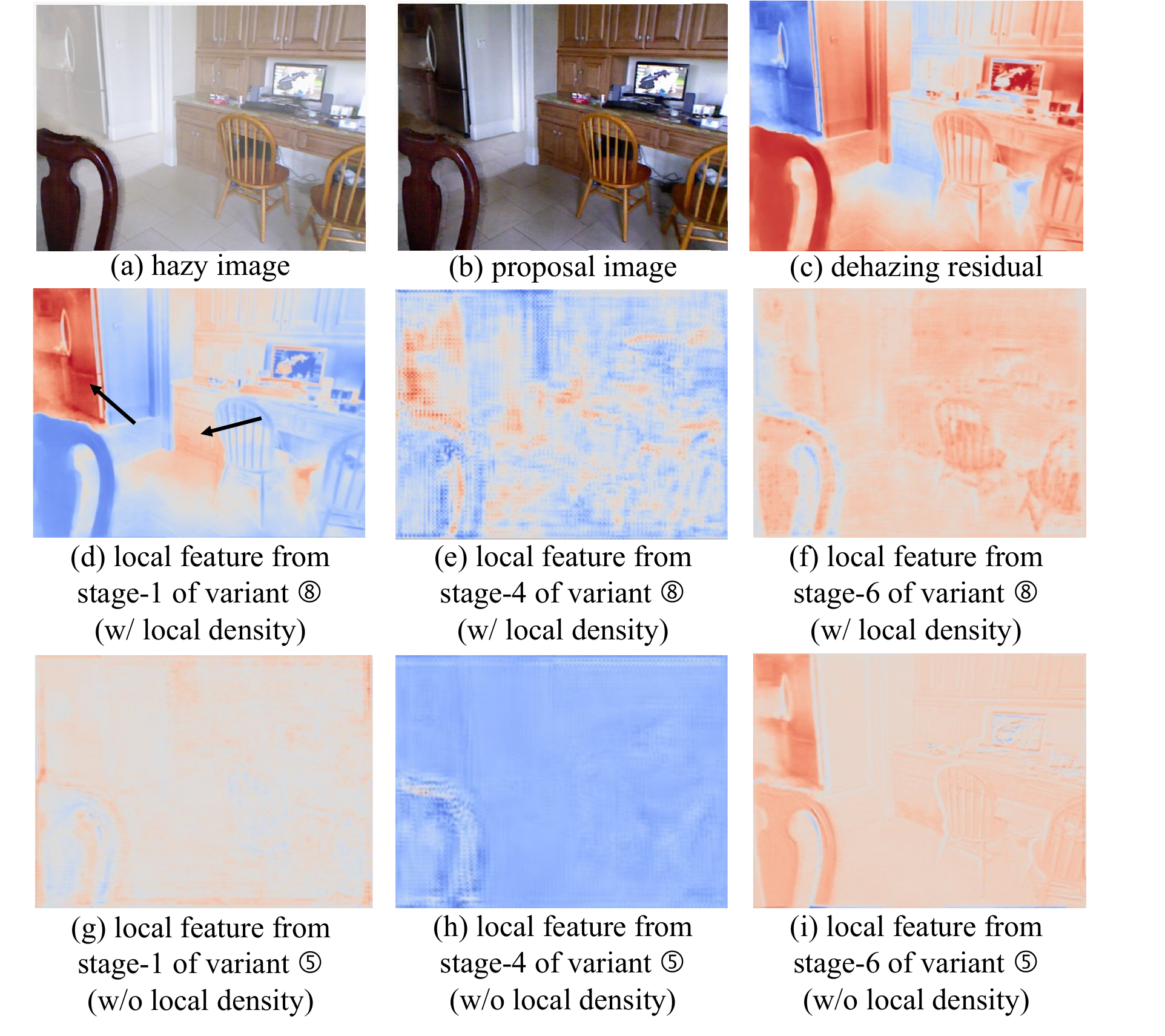}
  \caption{Visualization (heatmap form) of the local density features from variant \ding{179} and \ding{176}. The black arrows in (d) point out the hard regions found by the network by utilizing the dehazing residual.}
  \label{fig:ablation_local}
  
\end{figure}

\subsubsection{Ablation studies on utilizing global density difference and global feature refinement} We carry out experiments on variants \ding{173}, \ding{174}, \ding{175} to verify the effectiveness of our design using global density difference to extract and refine global density-related features. To ensure fairness, we adjust the ResBlock numbers of the Siamese structure to approximately match the number of parameters in the base network. Table \ref{tab:ablation} indicates that sharing parameters enables better feature learning from proposal images and delivers a 0.69 dB PSNR improvement. To motivate the network to learn more density-related information, $\mathcal{L}_{RD}$ is employed and the inclusion of it leads to a 0.34 dB PSNR improvement. In addition, the introduction of GDFR brings 1.10 dB PSNR performance gain by refining global densities features.

We also conducted experiments to validate the effectiveness of incorporating global density difference information. Fig. \ref{fig:ablation_global} showcases the dehazing results obtained without (variant \ding{172}) and with (variant \ding{175}) global density information, using image inputs with varying densities. When global density difference information is not introduced, we observed low quantitative metrics, unstable dehazing results, and inconsistent performance within the same scene. As depicted in Figure \ref{fig:ablation_global}, variant \ding{172} produces suboptimal dehazing outcomes.

However, upon incorporating global density differences, the network exhibits improved capability to perceive variations in global densities. This is achieved through the exploration of global density differences between the proposal image and the input image using the Siamese structure, as well as the refinement of density features using the Global Density Feature Refinement (GDFR) module. As a result, the network becomes more robust to images with different density levels, enabling the generation of consistent and visually clear dehazed images. 


\begin{table}[]
  \label{Table:ablation_IDRF}
\caption{
  Ablation studies on the usage of IDRF. \ding{52} indicates that the output features of this stage will be processed for IDRF.
}
\begin{tabular}{ccccccccc}
\toprule
\multirow{2}{*}{}     & \multicolumn{6}{c}{Stage} & \multirow{2}{*}{PSNR (dB)} & \multirow{2}{*}{Params (M)} \\ \cmidrule(l){2-7}
& 1  & 2  & 3  & 4  & 5 & 6 &  &  \\ \cmidrule(l){1-9} 
\multirow{3}{*}{IDRF} & \ding{52} & \ding{52} & \ding{52} &  &   &   &\underline{33.06}& 41.32 \\
&    &    &    & \ding{52} & \ding{52} & \ding{52} & 32.89 & 41.32 \\
& \ding{52} & \ding{52} & \ding{52} & \ding{52} & \ding{52} & \ding{52}  & \textbf{33.52} & 42.11 \\ \bottomrule
\end{tabular}
\end{table}

\subsubsection{Ablation studies on utilizing local density difference and local feature refinement} We perform experiments on variants \ding{177}, \ding{178}, \ding{179} for validation of the effectiveness of utilizing local density difference. As shown in Table \ref{tab:ablation}, extracting local features from dehazing residual, IDRF and $\mathcal{L}_{LDR}$ result in 0.12, 0.41 and 1.11 dB PSNR performance gains. 

To demonstrate the necessity of ultilizing local density difference, intermediate local features of the network using dehazing residual to extract features (variant \ding{179}) or not (variant \ding{176}) are visualized in Fig. \ref{fig:ablation_local}, which shows that high density and hard dehazing areas are captured in the shallow stage of our method. And as the network deepens, the local maps are gradually flattened, which implies the local features are updated stage by stage. However, variant \ding{176} fails to achieve these performances.

\subsubsection{Ablation studies on the location of IDRF usage}

We carried out ablation experiments to determine the optimal placement of the IDRF module within the network architecture. We compared three cases: using IDRF only in the encoder stages, only in the decoder stages, and in all stages. The results, as shown in Table \ref{Table:ablation_IDRF}, indicate that employing IDRF in all stages yields the best overall performance. 

Furthermore, we observed that incorporating IDRF in the encoder stage leads to improved quantitative metrics compared to its use in the decoder stages. This can be attributed to the fact that the refinement of local features should commence as early as possible in the network. The primary role of IDRF is to update the current local features and facilitate the alignment of the restored image features with clear image features. Without this refinement process, the network's local features remain relatively unchanged, resulting in less attention being paid to regions with relatively high density or hard features that require careful handling at the current stage. By introducing IDRF at an earlier stage, the network can be guided more effectively to bring the image features closer to the features of clear images.

\begin{table}[]
  \label{Table:ablation_DAFF}
\caption{
  Ablation studies on the effectiveness of DAFF. The experiments are conducted on the Haze4K dataset.
}
\begin{center}

\begin{tabular}{cccc}
\toprule
Methods &concat \cite{yeperceiving}& SK Fusion \cite{song2023vision}& DAFF\\ \cmidrule(l){1-4}
PSNR (dB) & 33.19 & \underline{33.80} & \textbf{34.63}\\
\bottomrule
\end{tabular}  
\end{center}
\label{tab:Ablation_DAFF}
\end{table}

\subsubsection{Ablation studies on the effectiveness of DAFF}
We further conduct ablation experiments on the effectiveness of DAFF. DAFF is designed to fully fuse the global density features passed by cross-branch connections, the image features of the LB branch, and the local density features. And we compare the PSNR performance with other two fusion strategies: concatenation \cite{yeperceiving} and SK Fusion \cite{song2023vision}. Table \ref{tab:Ablation_DAFF} demonstrate that our DAFF achieves better results than the other two methods. 

As described in Section \ref{sec:method}, DAFF first combines the global features with the image features, enabling the features to be aware of global density information. It then introduces the local features through the CSDA mechanism, facilitating pixel-level refinement of the features. In contrast, both concatenation and SK Fusion methods focus primarily on channel attention, without considering the attentional role of the local density map on the spatial dimensions of the features. The superior performance of DAFF can be attributed to its comprehensive fusion strategy, which takes into account both global and local density information. This enables our method to capture and utilize density-related information more effectively, resulting in improved dehazing performance compared to concatenation and SK Fusion methods.

\section{Conclusion}

In this paper, we propose DFR-Net, a density-aware method for image dehazing that utilizes haze density differences to extract and refine density-related features. To achieve this, we generate a proposal image and explore density representation from it and the hazy input. We use two branches to extract and refine global and local density features, respectively, based on the density differences between the proposal image and the hazy input. In Global Branch, features of images with different densities are pushed away, while in Local Branch, hazy image features are gradually pulled closer to those of clear images. Our experimental results demonstrate that DFR-Net achieves high-performance image dehazing ability, and our ablation studies show that our designs enable fine awareness and refinement of density information.


{\small
\bibliographystyle{unsrt}
\bibliography{egbib}
}



\end{document}